\documentclass{article}
\usepackage{ijcai24}
\pdfoutput=1

\usepackage{times}
\usepackage{soul}
\usepackage{url}
\usepackage{hyperref}
\usepackage[utf8]{inputenc}
\usepackage[small]{caption}
\usepackage{graphicx}
\usepackage{amsmath}
\usepackage{amsthm}
\usepackage{booktabs}
\usepackage{algorithm}
\usepackage{algorithmic}
\usepackage[switch]{lineno}

\usepackage{caption}
\usepackage{booktabs}
\usepackage{cellspace}
\usepackage{pifont}
\usepackage{amssymb}
\usepackage{tabularx}
\usepackage{xspace}
\usepackage{yhmath}
\usepackage{pifont}
\usepackage{adjustbox}
\usepackage{multirow}
\usepackage{makecell}
\usepackage{subfigure}
\usepackage{graphicx}
\usepackage{color}
\RequirePackage{CJKnumb}
\newcommand{\method}{\textsc{RMFNet}\xspace}

\title{Efficient Event Stream Super-Resolution with Recursive Multi-Branch Fusion}

\author{
Quanmin Liang$^{1,2*}$\and
Zhilin Huang$^{2,3*}$\and
Xiawu Zheng$^2$\and
Feidiao Yang$^2$\and\\
Jun Peng$^2$\and
Kai Huang$^1$\footnotemark[2]\And
Yonghong Tian$^{2,4}$\footnotemark[2]\quad\\
\affiliations
$^1$School of Computer Science and Engineering, Sun Yat-Sen University\\
$^2$Peng Cheng Laboratory\\
$^3$Shenzhen International Graduate School, Tsinghua University\\
$^4$Peking University\\
\emails
liangqm5@mail2.sysu.edu.cn,
\{zerinhwang03, yhtian\}@pku.edu.cn,
zhengxiawu@xmu.edu.cn,\\
\{yangfd, pengj01\}@pcl.ac.cn,
huangk36@mail.sysu.edu.cn
}
\usepackage{amsmath,amsfonts,bm}

\def\eqref#1{equation~\ref{#1}}
\def\1{\bm{1}}

\def\rvf{{\mathbf{f}}}

\def\rvn{{\mathbf{n}}}

\def\rvp{{\mathbf{p}}}

\def\rmA{{\mathbf{A}}}

\def\rmF{{\mathbf{F}}}

\def\rmK{{\mathbf{K}}}

\def\rmQ{{\mathbf{Q}}}

\def\rmV{{\mathbf{V}}}

\def\vf{{\bm{f}}}

\def\mC{{\bm{C}}}

\def\mR{{\bm{R}}}

\DeclareMathAlphabet{\mathsfit}{\encodingdefault}{\sfdefault}{m}{sl}
\SetMathAlphabet{\mathsfit}{bold}{\encodingdefault}{\sfdefault}{bx}{n}

\def\gE{{\mathcal{E}}}

\newcommand{\R}{\mathbb{R}}

\begin{document}

\maketitle

\begin{abstract}
    % 200个单词以内
    %Event Stream Super-Resolution (ESR) is a crucial technique for enhancing the resolution of event cameras. 
    Current Event Stream Super-Resolution (ESR) methods overlook the redundant and complementary information present in positive and negative events within the event stream, employing a direct mixing approach for super-resolution, which may lead to detail loss and inefficiency. 
    To address these issues, we propose an efficient Recursive Multi-Branch Information Fusion Network (\method) that separates positive and negative events for complementary information extraction, followed by mutual supplementation and refinement. 
    Particularly, we introduce Feature Fusion Modules (FFM) and Feature Exchange Modules (FEM). FFM is designed for the fusion of contextual information within neighboring event streams, leveraging the coupling relationship between positive and negative events to alleviate the misleading of noises in the respective branches. FEM efficiently promotes the fusion and exchange of information between positive and negative branches, enabling superior local information enhancement and global information complementation. 
    %To achieve this, we introduce attention-based feature fusion modules (FFM) and feature exchange modules (FEM), dedicated to integrating contextual information within neighboring event streams and the fusion-and-exchange of information from positive and negative branches, facilitating  superior local information enhancement and global information complementation. 
    %Additionally, we pioneer the exploration of the impact of data augmentation on ESR tasks and propose an effective data augmentation strategy that significantly enhances ESR performance and robustness. 
    Experimental results demonstrate that our approach achieves over \textbf{17\%} and \textbf{31\%} improvement on synthetic and real datasets, accompanied by a \textbf{2.3$\times$} acceleration. Furthermore, we evaluate our method on two downstream event-driven applications, \emph{i.e.}, object recognition and video reconstruction, achieving remarkable results that outperform existing methods. Our code and Supplementary Material are available at \href{https://github.com/Lqm26/RMFNet}{https://github.com/Lqm26/RMFNet}.
    %Our code will be released later.
\end{abstract}
\section{Introduction}
Event cameras are biologically inspired asynchronous sensors \cite{brandli2014240}. Unlike traditional cameras, event cameras register only the changes in brightness for each pixel over time. These are known as ``events'', which are categorized as positive or negative, depending on whether the brightness increases or decreases, respectively. This characteristic significantly reduces the amount of recorded information, resulting in advantages such as high temporal resolution, low power consumption, and a high dynamic range (HDR) \cite{gallego2020event}. However, as the application scenarios become more complex, the spatial resolution of existing event cameras is insufficient \cite{li2021event}. Increasing spatial resolution at the hardware level presents challenges in implementing asynchronous circuits \cite{gallego2020event}, making it difficult to maintain the low power consumption and high temporal resolution advantages of event cameras \cite{weng2022boosting,gehrig2022high}. %Therefore, some researchers propose to address this issue at the algorithm level.
Therefore, some researchers propose to address this issue at the software level, e.g. by leveraging advanced algorithms, which is referred to as Event Stream Super-Resolution (ESR).

\begin{figure}[!tp]
\centering
\includegraphics[width=\linewidth]{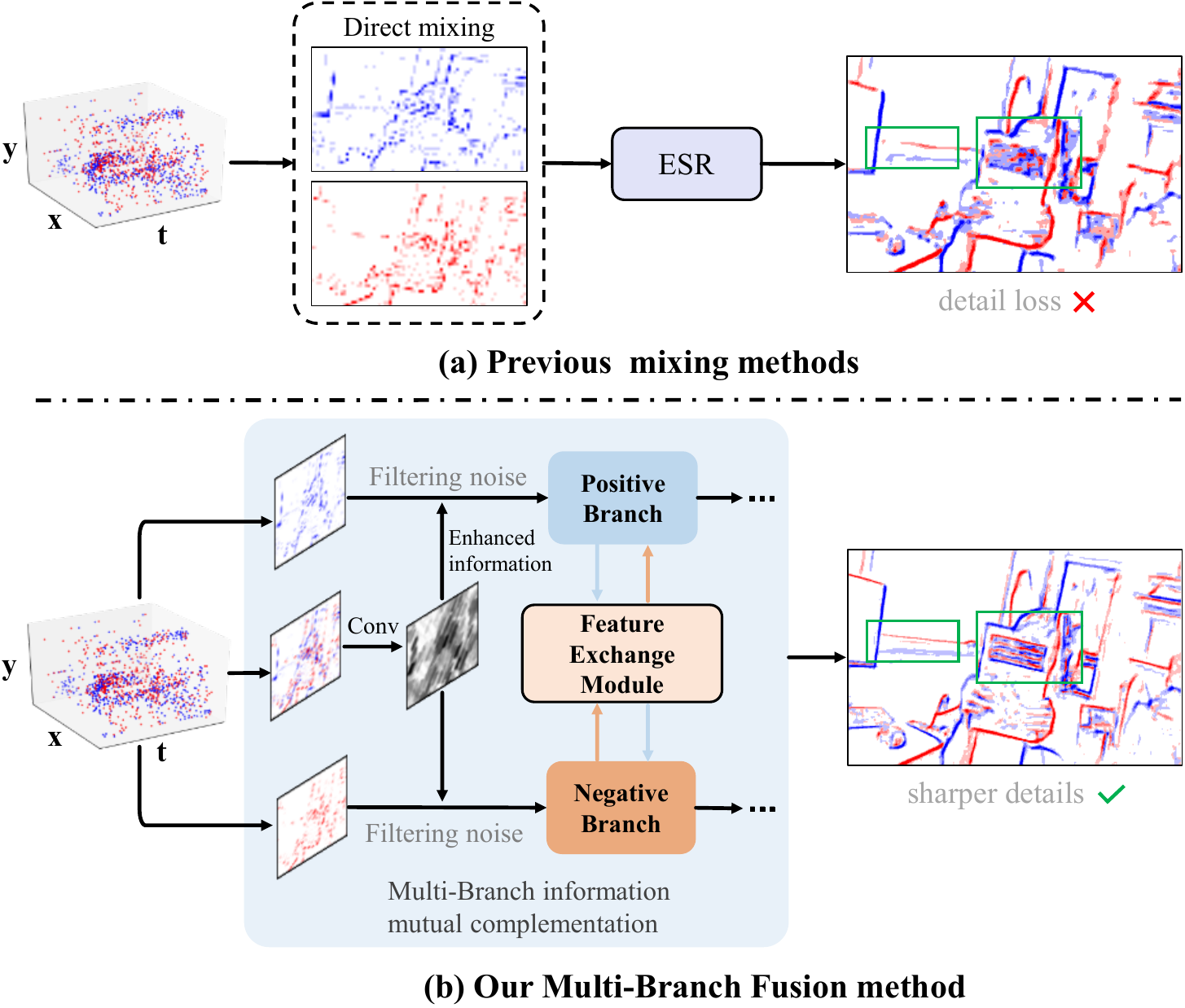}
\caption{Compared to previous ESR methods that directly mix positive and negative events, our multi-branch approach effectively extracts and integrates features from positive and negative events, achieving a more complete and clearer details (see the \textcolor{green}{green box}).}
\label{fig:motivation}
\end{figure}

Current research on ESR can be mainly divided into two directions. One approach aims to directly generate high-resolution event data from low-resolution event streams by spiking neural networks~\cite{li2019super,li2021event} or frame-assisted methods~\cite{wang2020joint}. However, these methods often require significant memory ~\cite{li2019super,li2021event} and high-quality images as assistance \cite{wang2020joint}, which complicates the training process and hinders achieving large-factor super-resolution.
Hence, researchers have proposed stacking event streams into either event frames \cite{rebecq2017real} or event count images \cite{maqueda2018event,zhu2018ev} and subsequently applying learning-based methods for ESR \cite{duan2021eventzoom,weng2022boosting}. Within event streams, there exist spatiotemporally inconsistent positive and negative events \cite{gehrig2019end}. These events do not perfectly align on a 2D plane but contain complementary information. Merging them into an event frame results in partial cancellation between positive and negative events, forming a new representation. As positive and negative events typically do not occur independently, the event frame helps filter out some naturally occurring noise in the event stream. Consequently, positive events, negative events, and event frames each contain different information about the event stream. However, previous methods did not effectively distinguish and fully utilize this information. They simply mix them and input them into the ESR model, leading to the loss of fine details in the super-resolved (SR) event stream (Figure~\ref{fig:motivation}(a)).

To address these issues, we propose an efficient Recursive Multi-Branch Information Fusion Network (\method). As illustrated in Figure~\ref{fig:motivation}(b), this network processes positive events, negative events, and event frames in a multi-branch fashion. Positive and negative events contain the majority of information in the event stream, while the event frame provide the guidance to filter noises. Therefore, we design a Feature Fusion Module (FFM) to highlight the valuable information in positive and negative streams according to the event frame at the initial stage. Specifically, this module calculates attention weight maps from features of different branches, facilitating the fusion of contextual information and aiding the positive and negative branches in noise removal. Subsequently, the positive and negative branches conduct feature extraction for positive and negative events, respectively, employing a Feature Exchange Module (FEM) for the adaptive fusion and exchange. Through capturing complementary information and long-range dependencies between positive and negative events, FEM improves the integration and exchange of information across different branches.

%Additionally, to further enhance the model's performance, we for the first time explore the impact of existing data augmentation methods on ESR tasks. We propose an effective data augmentation strategy that significantly improves the performance of ESR tasks.

The main contributions of our work are as follows:

\begin{itemize}
    \item We introduce an efficient Recursive Multi-Branch Information Fusion Network capable of effectively merging positive events, negative events, and event frames, thereby obtaining high-quality SR event images.
    \item We design Feature Fusion Modules and Feature Exchange Modules, which enhance the positive and negative event streams while effectively fusing and complementing information across different branches.
    \item We explored the impact of existing data augmentation methods on ESR tasks and proposed an effective data augmentation strategy to enhance the model's robustness and performance.
    \item Our method achieves over 17\% and 31\% improvement on synthetic and real datasets, accompanied by a 2.3$\times$ acceleration. In downstream event-based recognition and reconstruction tasks, our method effectively enhances performance, further validating the effectiveness of our approach.
\end{itemize}

\section{Related Work}
\textbf{Event Stream Super-resolution. }Due to the unique spatio-temporal characteristics of event streams, event stream super-resolution (ESR) tasks are often more challenging. Initially, Li et al. \cite{li2019super} introduced the Event Count Map (ECM) as a method to describe event spatial distribution. They established a spatiotemporal filter to generate a time-rate function and employed a non-homogeneous Poisson distribution to model events on each pixel. However, this approach encounters inaccuracies in estimating spatiotemporal distributions when performing high-factor super-resolution. To address this issue, Wang et al. \cite{wang2020joint} proposed a novel optimization framework called GEF, which utilizes motion correlation probabilities to filter event noise. The optimization maximizes the structural correspondence between low-resolution events and high-resolution image signals, facilitating event stream super-resolution in conjunction with image frames. Despite performing well in certain scenarios, the GEF method exhibits performance degradation when image frame quality deteriorates. Building upon this, Li et al. \cite{li2021event} proposed a spatio-temporal constraint learning method based on the spiking neural network (SNN) characteristics to simultaneously learn temporal and spatial features in event streams. On the other hand, Duan et al. \cite{duan2021eventzoom} transformed event streams into a 2D event frames format and designed a 3D U-Net-based network for ESR. While both methods demonstrated excellent performance in small-scale super-resolution tasks, they faced challenges of excessive memory requirements and training difficulties in large-factor super-resolution. To effectively address the challenges of large-factor super-resolution, Weng et al. \cite{weng2022boosting} introduced an event-based super-resolution method based on Recurrent Neural Networks. They initially transformed event streams into coarse-grained high-resolution event streams using coordinate relocation, followed by super-resolution through recurrent networks. This approach not only handles high-factor super-resolution effectively but also mitigates training challenges posed by excessive memory requirements.

However, these methods did not account for the spatiotemporal inconsistencies and complementarities between positive and negative events in the event stream. Directly mixing them may lead to the loss of details. Therefore, we propose \method, employing a multi-branch approach to mutually fuse and complement positive and negative events, which effectively enhances the performance of ESR.

% \section{Methods}
\section{Method}
\begin{figure*}[ht]
\centering
\includegraphics[width=\linewidth]{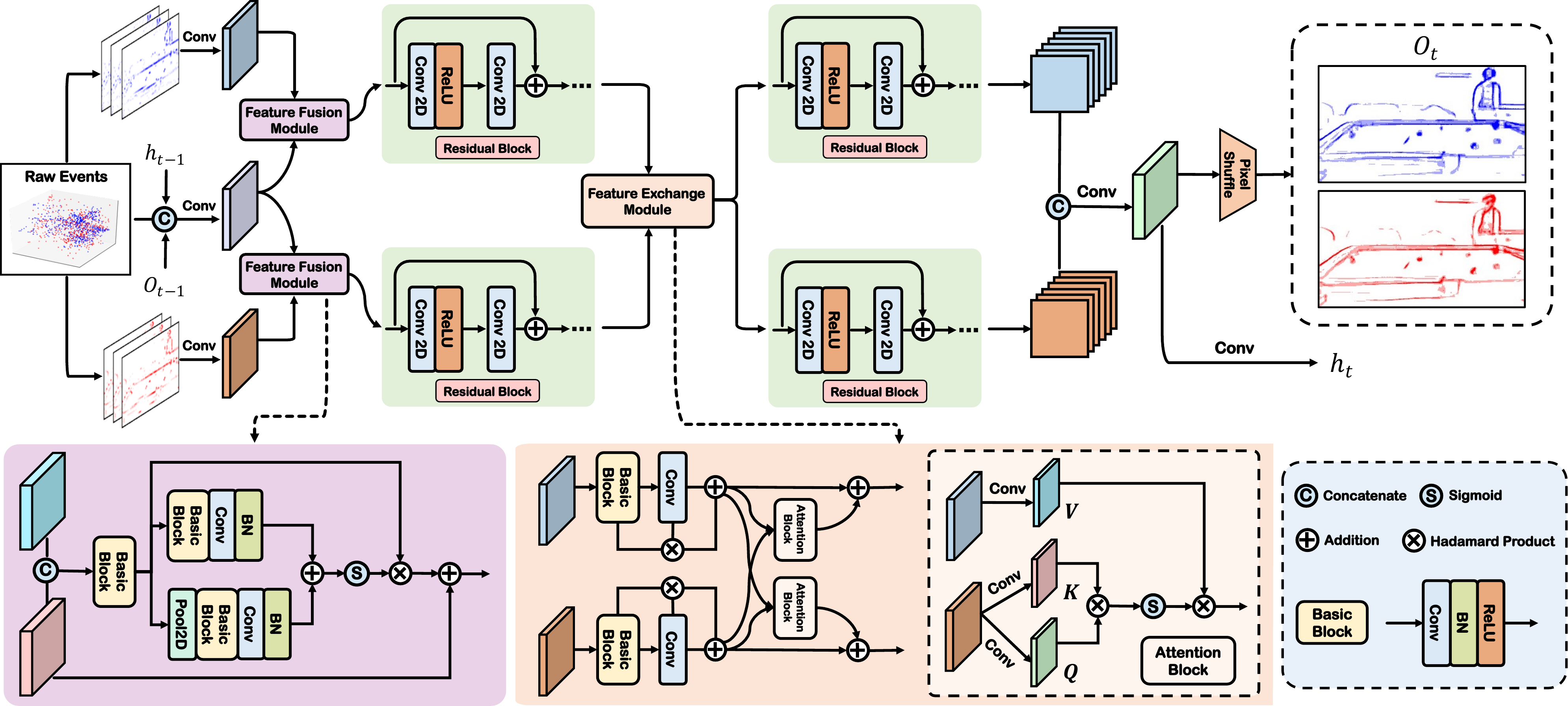}
\caption{Architecture of our proposed Recursive Multi-Branch Information Fusion Network (\method). Initially, the event frame is fused into positive and negative branches along with the previous output \(O_{t-1}\) and state \(h_{t-1}\) using the Feature Fusion Module (\textbf{bottom left}). Subsequently, each branch independently extracts features through Residual Blocks, and a Feature Exchange Module (\textbf{bottom}) facilitates the exchange of information between the branches. Finally, the features from the positive and negative branches are concatenated, and high-resolution event count images \(O_t\) are obtained through Pixel Shuffle operation.}
\label{fig:arch}
\end{figure*}
%Unlike the super-resolution methods for standard RGB images, event stream super-resolution methods typically involve three steps \cite{weng2022boosting,duan2021eventzoom}. Firstly, low-resolution event streams are stacked into a 2D representation, reducing the temporal dimension. Next, a super-resolution network is used to obtain a high-resolution 2D event image. Lastly, the temporal dimension of the event image is restored through resampling to obtain a high-resolution event stream.
%As depicted in Figure \ref{fig:arch}, we first stack the event stream into an event count image (including positive and negative events) and an event frame (Section \ref{data}). Subsequently, these are input into our \method (Section \ref{model}), which is a recursive network following the principles of recurrent neural networks~\cite{schuster1997bidirectional}. At each iteration, the input is augmented with the previous state. Within the \method, we introduce a Feature Fusion Module (Section \ref{FFM}) and a Feature Exchange Module (Section \ref{FEM}), dedicated to the fusion of contextual information within adjacent event streams and the exchange of information between positive and negative branches, respectively. Finally, the information from both branches is aggregated to produce a high-resolution event count image and a new state.
In this section, we first introduced the data representation methods for event cameras in Section \ref{data}. Subsequently, we presented the proposed Recursive Multi-Branch Information Fusion Network in Section \ref{model}. Finally, we described the data augmentation methods for ESR in Section \ref{DA}.

\subsection{Event Data Representation}
\label{data}
A set of event streams can be represented as $\gE = \{e_k\}^{N}_{k=1}$, where $N$ is the number of events, each event \(e_k \in \gE\) can be denoted by a tuple \((x_i, y_i, t_i, p_i)\), representing spatial coordinates, timestamp and polarity respectively. Subsequently, we partition $\{e_k\}^{N}_{k=1}$ into positive events $\{e_k\}^{N_p}_{k=1}$ and negative events $\{e_k\}^{N_n}_{k=1}$ based on their polarity \(p_i = \pm1\). Specifically, we stack $\{e_k\}^{N_p}_{k=1}$ and $\{e_k\}^{N_n}_{k=1}$ into event count images~\cite{maqueda2018event,zhu2018ev} according to the following equations:
\begin{equation}\label{eq1}
h\left( {x,y} \right) = {\sum\limits_{e_{k} \in \gE}{\delta\left( x - x_{i},~y - y_{i} \right)}}
\end{equation}
where \(\delta\) represents the Kronecker delta. Thus, we can build up two event count images from $\{e_k\}^{N_p}_{k=1}$ and $\{e_k\}^{N_n}_{k=1}$: positive $\rvp_t \in \R^{H\times W}$, and negative $\rvn_t \in \R^{H\times W}$. And the event frame~\cite{rebecq2017real} is obtained by stacking all events (including positive and negative events)  using equation (\ref{eq1}), resulting in $\rvf_t \in \R^{H\times W}$.
%Therefore, our objective is to perform super-resolution (SR) on the three event images of low-resolution (LR) event streams using our \method, resulting in high-resolution (HR) event images. Subsequently, these HR event images are resampled to obtain the final HR event stream.

\subsection{Multi-Branch Fusion Networks}
\label{model}
The framework of our proposed \method is depicted in Figure \ref{fig:arch}. The main inputs of this network include positive events, negative events, and event frames. Additionally, following a recursive approach \cite{schuster1997bidirectional}, we introduce the previous output \(O_{t-1}\) and state \(h_{t-1}\) into the input, aiding in better capturing features from adjacent event streams and achieving contextual fusion of event stream information. Given that positive and negative events contain the majority of information in the event stream, we process them separately through dedicated positive and negative branches. The event frame, serving as a coupled representation of positive and negative events, assists in filtering out noise (as positive and negative events do not occur independently). Thus, in the initial stage of \method, we fuse \(\rvf_t\) with \(O_{t-1}\) and \(h_{t-1}\) as event-enhanced information \(F_{Enh} \in \R^{C\times H\times W}\), which is passed to the positive and negative branches through the Feature Fusion Module (FFM). The positive and negative branches utilize Residual Block \cite{he2016deep} as the backbone for feature extraction from the fused positive and negative events, respectively. Subsequently, a Feature Exchange Module (FEM) is employed to facilitate the fusion and exchange of information between the positive and negative branches. Finally, the features from positive and negative events are concatenated, outputting the hidden state \(h_{t}\), and the SR event count images \(O_{t}\) are obtained through pixel shuffle \cite{shi2016real}.

\subsubsection{Feature Fusion Module}
\label{FFM}
As depicted in the bottom left corner of Figure \ref{fig:arch}, the FFM is tasked with transmitting event-enhanced information \(F_{Enh}\) to the positive and negative branches without compromising their distinctive features. Denoting the features extracted by convolutional layers for positive and negative events as \(F^{p}_{t} \in \R^{C\times H\times W}\) and \(F^{n}_{t} \in \R^{C\times H\times W}\), respectively.
%we exemplify the process using the positive branch. Our network is symmetric with respect to positive and negative branches.
Initially, we concatenate \(F^{p}_{t}\) with \(F_{Enh}\), followed by a preliminary fusion through a Basic Block, resulting in \(F^{fuse}_t\). The event-enhanced information encompasses details from adjacent event streams and coupling information between positive and negative events, effectively guiding the positive branch in detail recovery. To integrate these features seamlessly, we utilize the fused feature \(F^{fuse}_t\) to compute two attention weights. The first is local attention weight:
\begin{equation}\label{eq2}
\rmA_{t}^{loc} = BN\left( \mC_{1\times1}\left( {\mR\left( {BN\left( {\mC_{1\times1}\left( F_{t}^{fuse} \right)} \right)} \right)} \right) \right)
\end{equation}
where \(BN\) denotes batch normalization, \(\mC_{1\times1}\) represents a \(1\times1\) convolution operation, and \(\mR\) represents the ReLU activation function.

The second is global attention weight \(\rmA_{t}^{glo} \in \R^{C\times 1\times 1} \), which is computed channel-wise. Specifically, we incorporate global average pooling to process \(F^{fuse}_t\) along the spatial dimensions:
%The second attention weight is computed globally in a channel-wise. Hence, we incorporate global average pooling to process \(F^{fuse}_t\) along the spatial dimensions, utilizing the following equation:
\begin{align}\label{eq3}
\rmA_{t}^{glo} = \vf^{att}\left( GAP\left( F_{t}^{fuse} \right) \right)
\end{align}
where \(\vf^{att}\) represents the function given in equation (\ref{eq2}), and \(GAP\) stands for global average pooling.
%and \(\rmA_{t}^{glo} \in \R^{C\times 1\times 1}\).

Finally, we combine the global and local attention, apply it to \(F^{fuse}_t\), and add it to the previous features of positive events \(F^{p}_{t}\), thereby integrating the event-enhanced information into the positive branch:
\begin{equation}\label{eq4}
F_{t}^{out} = F_{t}^{p} + F_{t}^{fuse}\otimes\left( \sigma\left( \rmA_{t}^{glo} \oplus \rmA_{t}^{loc} \right) \right)
\end{equation}
where \(\otimes\) represents element-wise product, \(\sigma\) denotes the sigmoid activation function, and \(\oplus\) signifies broadcasting addition. Our network is entirely symmetric with respect to positive and negative branches, so the negative branch follows the same process.

\subsubsection{Feature Exchange Module}
\label{FEM}
Considering the complementary and redundant information present in positive and negative events, directly integrating feature information from these branches may have adverse effects. To address this, we introduce a Feature Exchange Module (depicted below Figure \ref{fig:arch}), which utilizes attention mechanisms to automatically select and enhance crucial features, facilitating efficient information exchange between the two branches.

Firstly, to reduce the redundancy in individual branch features and emphasize important features, we apply spatial attention separately to both branches:
\begin{equation}\label{eq5}
\Tilde\rmF_{t} = {Conv}_{basic}\left( F_{t}^{in} \right)
\end{equation}
\begin{equation}\label{eq6}
\Tilde\rmF_{t}^{P} = Conv\left( \Tilde\rmF_{t} \right)\otimes\Tilde\rmF_{t} + Conv\left( \Tilde\rmF_{t} \right)
\end{equation}
where \(F_{t}^{in}\) represents the input features from the positive and negative branches, \({Conv}_{basic}\) denotes the Basic Block, and \(Conv\) represents the convolutional operation. The \(Conv(\Tilde\rmF_{t})\) in Eq.(\ref{eq6}) respectively serves as the weight and bias, adjusting the weights of branch features. \(\Tilde\rmF_{t}^{P}\) is the output of the positive branch, and \(\Tilde\rmF_{t}^{N}\) is obtained similarly from the negative branch.

Subsequently, inspired by self-attention mechanisms \cite{vaswani2017attention,wang2018non}, we design two symmetrical Attention Blocks to capture complementary information from the positive and negative branches. Taking the positive branch as an example, we use \(\mC_{1\times1}\) to obtain \(\rmV \in \R^{C\times (HW)}\) for the positive branch, and \(\rmQ \in \R^{C_1\times (HW)}\) and \(\rmK \in \R^{C_1\times (HW)}\) for the negative branch. Here, \(C\) represents the number of channels, and \(C_1\) is set to 1/8 of \(C\) for enhanced computational efficiency. Therefore, the output of the positive branch, fused with features from the negative branch, can be represented as:
\begin{equation}\label{eq7}
\rmF_{fuse}^{P} = \rmV\otimes\left( \sigma\left( \rmQ^\mathsf{T}\otimes\rmK \right) \right)
\end{equation}
where \(\rmQ^\mathsf{T}\) represents the transpose of \(\rmQ\). Through the two symmetrical attention modules, we can achieve a complementary fusion of features from the positive and negative branches, effectively enhancing the performance of ESR.

\subsubsection{Training Objectives}
We partition the event stream into multiple sequences of length \(T\) for training our method, following the approach of Weng et al \cite{weng2022boosting}. We set \(T=9\) and use Mean Squared Error (MSE) to calculate the loss:
\begin{equation}\label{eq8}
\mathcal{L} = ~{\sum_{t = 1}^{T}{MSE\left( {O}_{t}^{SR},~{ECI}_{t}^{HR} \right)}}
\end{equation}
where \({O}_{t}^{SR}\) represents the event count images of the final SR event stream, \({ECI}_{t}^{HR}\) represents the ground truth event count images, and \(MSE\) is the mean square error function.

\subsection{Data Augmentation for ESR}
\label{DA}
Previous research in the field of image or video super-resolution has shown that methods involving operations or augmentations in the pixel space \cite{zhang2017mixup,yun2019cutmix} can effectively enhance task performance \cite{yoo2020rethinking}, as they preserve the spatial relationships within the images. In the realm of ESR, there is currently a lack of systematic investigation into Event Stream Super-Resolution Data Augmentation (ESRDA). To address this gap, we adapt and refine data augmentation methods from some event stream studies \cite{gu2021eventdrop,barchid2023exploring} and RGB image domains, exploring the impact of data augmentation on the ESR task. We experiment with the following methods:
\begin{itemize}
    \item \textbf{Polarity flipping}.
    % : Inverts the polarity \(p_i\) of event stream.
    \item \textbf{RandomFlip} \cite{SimonyanZ14a}.
    % : Applies horizontal or vertical flipping to event images.
    \item \textbf{Drop by time} \cite{gu2021eventdrop}.
    % : Discards a portion of the same time interval in both Low-Resolution (LR) and High-Resolution (HR) event streams.
    \item \textbf{Random drop} \cite{gu2021eventdrop}.
    % : Randomly drops events in the LR event stream based on a certain proportion.
    \item \textbf{Drop by area} \cite{gu2021eventdrop}.
    % : Similar to the CutOut operation \cite{devries2017improved}, simultaneously discards a region in both LR and HR event images based on a given proportion.
    \item \textbf{Random drop or add noise}.
    % : We propose randomly dropping events in the LR event stream or introducing noisy events based on a certain proportion.
    \item \textbf{Static Translation}.
% : Horizontally or vertically shifts both LR and HR event images by a certain proportion.
    \item \textbf{RandomResizedCrop} \cite{he2016deep}.
    % : Applies random cropping to LR and HR event images, restoring them to the original size through nearest interpolation.
\end{itemize}
Regarding the details and parameters for data augmentation operations, please refer to the \textbf{Supplementary Material}.

\begin{figure*}[ht]
  \centering
  \includegraphics[width=\textwidth]{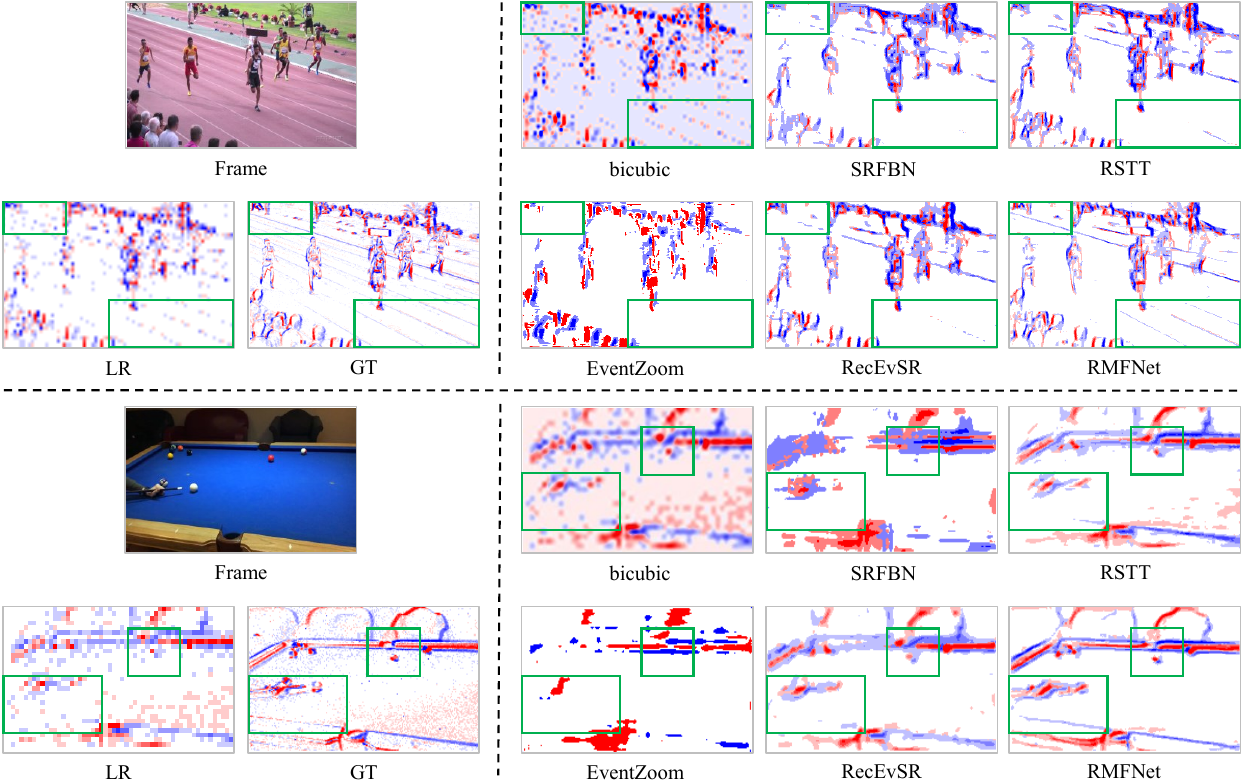}
  \caption{Qualitative analysis comparison on synthetic and real datasets. The upper and lower figures represent \(4\times\) SR results on the NFS-syn and EventNFS datasets, respectively. It is evident that our \method excels in recovering finer details of the event streams on both datasets (see the \textcolor{green}{green box}), resulting in sharper edges. Positive events are in \textcolor{blue}{blue}, negative events in \textcolor{red}{red}. \textbf{Zoom in for the best view}.}
  \label{figure2}
\end{figure*}
\begin{table*}
\centering
\begin{adjustbox}{width=0.95\textwidth}
    \renewcommand{\arraystretch}{1.2}
\begin{tabular}{lccccccccccccccccc}
\toprule
\multirow{2}{*}{Methods} &
  \multicolumn{3}{c}{NFS-syn} &
   &
  \multicolumn{2}{c}{RGB-syn} &
   &
  \multicolumn{2}{c}{EventNFS-real} &
   &
  \multicolumn{3}{c}{Param (M)} &
   &
  \multicolumn{3}{c}{Inference time (ms)} \\ \cmidrule{2-4} \cmidrule{6-7} \cmidrule{9-10} \cmidrule{12-14} \cmidrule{16-18} 
          & \(2\times\)    & \(4\times\)    & \(8\times\)    &  & \(2\times\)    & \(4\times\)     &  & \(2\times\)    & \(4\times\)    & & \(2\times\)    & \(4\times\)    & \(8\times\)   &  & \(2\times\)    & \(4\times\)    & \(8\times\)    \\ \midrule
bicubic   & 0.616 & 0.531 & 0.545 &  & 0.1197 & 0.1429 &  & 0.760 & 0.899 &  & -    & -    & -    &  & -    & -    & -     \\
SRFBN     & 0.411 & 0.394 & 0.394 &  & 0.1051 & 0.1010 &  & 0.415 & 0.545 &  & \underline{2.1}  & 3.6  & 7.9  &  & 37.3 & 54.8 & 65.4  \\
RSTT     & \underline{0.389} & \underline{0.366} & 0.365 &  & \underline{0.0954} & \underline{0.0909} &  & \underline{0.310}  & \underline{0.399} &  & 3.8  & 4.1  & 4.3  &  & 61.4 & 61.1 & 73.0    \\ \midrule
EventZoom & 0.806 & 1.049 & 1.239 &  & 0.4462 & 1.2232 &  & 0.778 & 1.248 &  & 11.5 & 11.5 & 11.5 &  & 17.4 & 70.1 & 396.6 \\
RecEvSR   & 0.430  & 0.368 & \underline{0.332} &  & 0.5013 & 0.3360 &  & 0.376 & 0.449 &  & \textbf{1.8}  & \textbf{1.8}  & \textbf{1.8}  &  & \underline{13.2} & \underline{18.9} & \underline{19.2}  \\ \midrule
Ours      & \textbf{0.316} & \textbf{0.300} & \textbf{0.305} &  & \textbf{0.0899}  & \textbf{0.0865} &  & \textbf{0.250}  & \textbf{0.316} &  & 3.0  & \underline{3.1}  & \underline{3.6}  &  & \textbf{7.0}  & \textbf{7.5}  & \textbf{7.8}   \\ \bottomrule
\end{tabular}
\renewcommand{\arraystretch}{1}
\end{adjustbox}
\caption{Quantitative analysis comparison on real and synthetic datasets. Mean Squared Error (\(MSE\)) is used as the evaluation metric. Model Parameters (Param) and Inference time are calculated on the NFS-syn dataset. \textbf{Bold} and \underline{underline} indicate the best and second-best results.}
\label{table1}
\end{table*}
\section{Experiments}
\subsection{Datasets and Training Settings}
Obtaining event data is challenging, and the availability of event datasets containing LR-HR pairs at multiple scales is limited. To address this scarcity, similar to many event-based tasks \cite{weng2022boosting,rebecq2019high,wang2020eventsr}, we employed synthetic simulation datasets to enrich our training data. EventNFS \cite{duan2021eventzoom} is the first dataset to include LR-HR pairs captured through a designed display-camera system, capturing rapidly displayed images on a monitor. However, due to device resolution limitations, the minimum resolution is \(55\times31\), and there are only \(4\times\) data pairs at the maximum. Moreover, the data at the smallest resolution suffers severe degradation due to its low resolution. To overcome these issues, we utilized an event simulator \cite{lin2022dvs} to transform the NFS dataset \cite{kiani2017need} and RGB-DAVIS dataset \cite{wang2020joint} into event data, resulting in NFS-syn and RGB-syn datasets. We selected these datasets because of their high temporal resolution, which can better simulate real-world event streams.
%Specifically, we downsampled the NFS dataset by \(2(4,8)\times\) and then generated event data using the event simulator, with resolutions ranging from \(640\times360\) to \(80\times45\). This dataset includes 100 sequences from 65 scenes. In the RGB-syn dataset, we constructed \(2(4)\times\) event pairs from 20 sequences, consisting of 10 indoor and 10 outdoor scenes. We applied data augmentation and random splitting for training and testing, following previous practices \cite{weng2022boosting, duan2021eventzoom}.
For further details, please refer to the \textbf{Supplementary Material}.

For a fair comparison, we maintained training settings consistent with \cite{weng2022boosting}. We used \(MSE\) as the evaluation metric for our models. All experiments were conducted on a Tesla V100 GPU.
%The initial learning rate was set to \(10^{-3}\), with a decay factor of 0.95 and decay applied every 4000 iterations. All models were trained for 100000 iterations, with a batch size of 2, and the entire set of experiments was conducted on an NVIDIA RTX 2080 Ti GPU.

\subsection{Comparison with State-of-the-Art Models}
In this work, we primarily compared our proposed \method with two previous learning-based approaches, EventZoom \cite{duan2021eventzoom} and RecEvSR \cite{weng2022boosting}. Other ESR methods \cite{li2021event,wang2020joint,li2019super} relying on real frames as assistance or prone to failure in complex scenes, pose challenges for fair comparisons. EventZoom, being the first learning-based event stream super-resolution method, faces challenges in training for large-scale SR due to its 3D-Unet architecture, making it difficult and computationally expensive. To address this, following previous practices \cite{weng2022boosting}, we ran EventZoom-\(2\times\) multiple times to obtain results for larger SR factors.
%RecEvSR stands as the current SOTA model for event stream super-resolution, employing a recurrent neural network design to tackle large-scale SR challenges. We utilized the code provided by the authors for model training.
Additionally, we include classic image super-resolution methods such as bicubic and SRFBN \cite{li2019feedback}, as well as a transformer-based video super-resolution method, RSTT \cite{geng2022rstt}, for comparison. We randomly split the real dataset EventNFS for training and testing. To evaluate the model's generalization, we select a subset of NFS-syn data for \(2(4,8)\times\) SR training and then validate on both the NFS-syn and RGB-syn datasets.
%To assess model generalization, we conducted \(2(4,8)\times\) SR training using a subset of NFS-syn data and validated on the EventNFS and RGB-syn datasets.

\textbf{Qualitative Analysis Results.} As depicted in Figure \ref{figure2}, we present the \(4\times\) SR results of various methods on both synthetic and real data (for more results, please refer to the \textbf{Supplementary Material}). It can be observed that traditional image super-resolution methods such as bicubic and SRFBN \cite{li2019feedback} struggle to achieve satisfactory visual results in ESR tasks, exhibiting blurry edges and significant detail loss. This may be attributed to the gap between event stream and RGB images. EventZoom \cite{duan2021eventzoom}, on the other hand, exhibits numerous detail losses, likely due to error accumulation from multiple runs of EventZoom-\(2\times\). In comparison, RSTT \cite{geng2022rstt} and RecEvSR \cite{weng2022boosting} produce event images of higher quality, yet they still fall short in detail restoration and supplementation. In contrast, our proposed \method can better extract detailed information from positive and negative event streams and complement each other, resulting in more comprehensive details and clearer edge information.

\textbf{Quantitative Analysis Results.} As shown in Table \ref{table1}, compared to the previously SOTA ESR method RecEvSR, \method achieves an average MSE improvement of 17.7\% and 31.6\% on NFS-syn and EventNFS, respectively. On RGB-syn, RecEvSR exhibits fragile generalization, while our \method maintains good generalization with an average MSE improvement of 78\%. Additionally, the average inference speed is improved by \(2.3\times\). Compared to the video super-resolution method RSTT, \method achieves an average MSE improvement of 17.8\% and 20\% on NFS-syn and EventNFS, respectively. On RGB-syn, while RSTT maintains good generalization, our \method still outperforms it with a 5.3\% improvement. Furthermore, our inference speed is improved by \(8.7\times\). These results demonstrate the efficiency and robustness of our proposed \method.

\begin{table}[ht]
    \renewcommand{\arraystretch}{1.2}
\begin{adjustbox}{width=1\linewidth}
\begin{tabular}{l|ccc}
\toprule
Method                   & NFS-syn & RGB-syn & EventNFS \\ \midrule
\method (w/o DA)            & 0.304   & 0.0874  & 0.795    \\ \midrule
\textbf{Polarity flipping}        & 0.304   & 0.0865 \textcolor{green}{\(\downarrow\)}  & 0.793 \textcolor{green}{\(\downarrow\)}    \\
\textbf{RandomFlip}               & 0.302 \textcolor{green}{\(\downarrow\)}   & 0.0868 \textcolor{green}{\(\downarrow\)}  & 0.793 \textcolor{green}{\(\downarrow\)}    \\
\textbf{Drop by time}             & 0.304   & 0.0873 \textcolor{green}{\(\downarrow\)}  & 0.790 \textcolor{green}{\(\downarrow\)}    \\
Random drop              & 0.306 \textcolor{red}{\(\uparrow\)}  & 0.0880 \textcolor{red}{\(\uparrow\)}  & 0.785 \textcolor{green}{\(\downarrow\)}    \\
Drop by area             & 0.306 \textcolor{red}{\(\uparrow\)}   & 0.0881 \textcolor{red}{\(\uparrow\)}  & 0.783 \textcolor{green}{\(\downarrow\)}    \\
\textbf{Random drop or add noise} & 0.303 \textcolor{green}{\(\downarrow\)}   & 0.0871 \textcolor{green}{\(\downarrow\)}  & 0.786 \textcolor{green}{\(\downarrow\)}    \\
Static Translation       & -       & -       & -        \\
RandomResizedCrop        & 0.330 \textcolor{red}{\(\uparrow\)}   & 0.0921 \textcolor{red}{\(\uparrow\)}  & 0.799 \textcolor{red}{\(\uparrow\)}    \\ \midrule
Selected DA's (random)    & \textbf{0.300} \textcolor{green}{\(\downarrow\)}   & \textbf{0.0865} \textcolor{green}{\(\downarrow\)}  & \textbf{0.771} \textcolor{green}{\(\downarrow\)}    \\ \bottomrule
\end{tabular}
\renewcommand{\arraystretch}{1}
\end{adjustbox}
\caption{Comparison of different data augmentation methods in ESR task. Training is conducted on the NFS-syn dataset, and \(4\times\) SR testing is performed on NFS-syn, RGB-syn, and EventNFS datasets.}
\label{table2}
\end{table}
\subsection{Analysis of ESRDA Methods}
As shown in Table \ref{table2}, we compared the impact of different DA methods on our \method for the \(4\times\) ESR task. To better highlight the influence of data augmentation methods on the generalization of our model, we only conducted training on NFS-syn and performed testing on NFS-syn, RGB-syn, and EventNFS. It can be observed that Polarity flipping, RandomFlip, and Drop by time all contribute to performance gains in the ESR task. However, Static translation leads to training instability, while RandomResizedCrop and Drop by area result in a decline in the performance and generalization of our \method. This suggests that altering the relative spatial relationships between events may adversely affect the ESR task. This could be attributed to the sparse and unidimensional nature of event streams, lacking important features such as color and intensity. Therefore, disrupting the relative spatial relationships among event streams significantly impacts the overall structure, introducing additional noise and consequently leading to a decline in model performance.

Random drop only discards a portion of events in the LR event stream, introducing potential biases in model fitting. To address this, we propose Random drop or add noise, where events are not only dropped with a certain probability but noise is also added simultaneously, mitigating this issue and enhancing model robustness. Lastly, inspired by RandAugment \cite{cubuk2020randaugment}, we combine Polarity flipping, RandomFlip, Drop by time, and Random drop or add noise into a data augmentation ensemble, from which one augmentation is randomly selected (Selected DA). Experimental results demonstrate that our DA strategy effectively enhances the performance and generalization of our model in the ESR task. For more related experiments, please refer to the \textbf{Supplementary Material}.

\begin{table}[ht]
\begin{adjustbox}{width=1\linewidth}
    \renewcommand{\arraystretch}{1.2}
\begin{tabular}{lccccc}
\toprule
Model & Multi-Branch & FFM & FEM & NFS-syn & EventNFS \\ \midrule
model\#A     & \ding{56}            & \ding{56}   & \ding{56}   & 0.329   & 0.347    \\
model\#B     & \ding{52}            & \ding{56}   & \ding{56}   & 0.317   & 0.331    \\
model\#C     & \ding{52}            & \ding{56}   & \ding{52}   & 0.309   & 0.322    \\
model\#D     & \ding{52}            & \ding{52}   & \ding{56}   & 0.313   & 0.326    \\
model\#E     & \ding{52}            & \ding{52}   & \ding{52}   & \textbf{0.300}   & \textbf{0.316}    \\ \bottomrule
\end{tabular}
\renewcommand{\arraystretch}{1}
\end{adjustbox}
\caption{Ablation results for different components of our \method.}
\label{table3}
\end{table}
\subsection{Ablation Study}
To validate the effectiveness of different components in our proposed \method, we conducted experiments with four different variants and compared the \(4\times\) SR results on the NFS-syn and EventNFS datasets.

As shown in Table \ref{table3}, we compared \method with several variants with different settings: 1) \textbf{model\#A}: using a single-branch model, concatenating event images and state \(h_t\) at the initial stage, and then inputting them into the model. 2) \textbf{model\#B}: discarding FFM and FEM modules, using lateral connections \cite{feichtenhofer2019slowfast,christoph2016spatiotemporal} between branches as an alternative. 3) \textbf{model\#C}: discarding the FFM module, using lateral connections between branches as an alternative. 4) \textbf{model\#D}: discarding the FEM module, using lateral connections between positive and negative branches as an alternative.

According to the results in Table \ref{table3}, the multi-branch model significantly outperforms the single-branch model, as it effectively decouples different parts of the event stream, allowing for fine-grained learning of each part's features. Additionally, the FFM and FEM designed in our model efficiently fuse and exchange features from different branches, promoting information complementarity between positive and negative event streams, outperforming methods that directly mix features from different branches. For more details about model hyperparameter ablation experiments, please refer to the \textbf{Supplementary Material}.

%\subsection{Model Analysis}
%We also compared RLSP, EventZoom, RecEvSR, and our MRN in terms of model parameters and inference time (Table 5). We conducted measurements on the NFS-syn dataset for \(2(4,8)\times\) SR tasks. It's evident that RLSP has the fewest parameters and significantly outperforms other methods in terms of inference time. However, as mentioned earlier, it may suffer from issues like detail loss in complex scenes. EventZoom, due to the need for multiple executions of \(2\times\) networks for achieving high-factor SR, experiences rapid increases in inference time. Despite RecEvSR having fewer parameters, its complex structural design leads to longer inference times. In contrast, our MRN, despite having slightly more parameters, achieves a good balance between parameter count and inference time. Its well-designed architecture allows for fast inference that is only marginally higher than RecEvSR's, while effectively restoring event stream details. Furthermore, as previously mentioned, we can optimize our \method to achieve satisfying parameter counts and inference times based on specific requirements.

\subsection{Event-based Applications}
\begin{table}[t]
\centering
\begin{adjustbox}{width=1\linewidth}
    \renewcommand{\arraystretch}{1.2}
\begin{tabular}{lllllllll}
\toprule
& \multicolumn{8}{c}{Video Reconstruction}              \\
\multirow{2}{*}{Methods} & \multicolumn{2}{c}{\(2\times\)}          &           & \multicolumn{2}{c}{\(4\times\)}          &           & \multicolumn{2}{c}{\(8\times\)}          \\ \cmidrule{2-3} \cmidrule{5-6} \cmidrule{8-9} 
                         & SSIM \(\uparrow\)  & LPIPS \(\downarrow\) &  & SSIM \(\uparrow\)  & LPIPS \(\downarrow\) &  & SSIM\(\uparrow\)  & LPIPS \(\downarrow\) \\ \midrule
bicubic                  & 0.568 & 0.395 &  & 0.609 & 0.522 &  & 0.598 & 0.545 \\
SRFBN                    & 0.608 & 0.389 &  & 0.618 & 0.455 &  & 0.612 & 0.489 \\
RSTT                     & \underline{0.627} & \underline{0.359} &  & \underline{0.639} & \underline{0.424} &  & 0.622 & 0.472 \\
EventZoom                & 0.542 & 0.429 &  & 0.575 & 0.488 &  & 0.574 & 0.542 \\
RecEvSR                  & 0.611 & 0.371 &  & 0.637 & 0.426 &  & \underline{0.630} & \underline{0.466} \\
\method & \textbf{0.648} & \textbf{0.339} &  & \textbf{0.667} & \textbf{0.409} &  & \textbf{0.653} & \textbf{0.450} \\ \bottomrule
\end{tabular}
\renewcommand{\arraystretch}{1}
\end{adjustbox}
\begin{adjustbox}{width=1\linewidth}
    \renewcommand{\arraystretch}{1.2}
\begin{tabular}{lllllllll}
\toprule
          \multirow{2}{*}{Methods} & \multicolumn{8}{c}{Object Recognition}            \\ \cmidrule{2-9} 
          & ACC \(\uparrow\)   & AUC \(\uparrow\)   &  & ACC \(\uparrow\)   & AUC \(\uparrow\)  &  & ACC \(\uparrow\)  & AUC \(\uparrow\)  \\ \midrule
bicubic   & 56.67 & 57.43 &  & 56.01 & 56.89 &  & 49.95 & 50.77 \\
SRFBN     & 61.12 & 61.94 &  & 60.89 & 61.03 &  & 50.02 & 50.86 \\
RSTT      & \underline{63.51} & \underline{63.96} &  & \underline{63.02} & \underline{64.29} &  & 52.97 & 54.07 \\
EventZoom & 54.68 & 56.03 &  & 49.56 & 50.45 &  & 47.96 & 48.74 \\
RecEvSR   & 62.91 & 63.47 &  & 62.37 & 63.07 &  & \underline{53.57} & \underline{54.48} \\
\method & \textbf{68.75} & \textbf{69.56} &  & \textbf{69.52} & \textbf{69.80} &  & \textbf{58.16} & \textbf{59.05} \\ \midrule
GT                       & 85.16 & 84.99 &  & 93.44 & 93.52 &  & 94.96 & 94.81 \\ \bottomrule
\end{tabular}
\renewcommand{\arraystretch}{1}
\end{adjustbox}
\caption{Quantitative comparison for event-based video reconstruction and object recognition. Video reconstruction is conducted on the NFS-syn dataset, while object recognition is performed on the NCars dataset \protect\cite{sironi2018hats}. AUC and ACC represent accuracy and area under the curve, respectively. GT denotes the result obtained by directly using downsampled event streams for recognition. \textbf{Bold} and \underline{underline} indicate the best and second-best results.}
\label{tab4}
\end{table}

\textbf{Video Reconstruction.} Video reconstruction is a crucial task within event-based applications \cite{rebecq2019high,stoffregen2020reducing,weng2021event,liang2023event,yang2023learning}. Firstly, we conducted \(2(4,8)\times\) SR on NFS-syn using bicubic, SRFBN \cite{li2019feedback}, RSTT \cite{geng2022rstt}, EventZoom \cite{duan2021eventzoom}, RecEvSR \cite{weng2022boosting}, and our \method. Subsequently, we adopt E2VID \cite{rebecq2019high} as the benchmark algorithm for event-based video reconstruction and utilize the structural similarity (SSIM) \cite{wang2004image} and the perceptual similarity (LPIPS) \cite{zhang2018unreasonable} as evaluation metrics for reconstruction quality. Table \ref{tab4} presents the quantitative results for event-based video reconstruction, indicating that our method surpasses others in both SSIM and LPIPS metrics, and exhibits more visually satisfying details (see \textbf{Supplementary Material}). This further underscores our method's capability to better restore details in the LR event stream.

\textbf{Object Recognition.} We also perform a comparison of all models and methods in the event-based object recognition task. In this context, following the methodology of Weng et al. \cite{weng2021event}, we employ the NCars dataset \cite{sironi2018hats} for experimentation and leverage the classifier proposed by Gehrig et al. \cite{gehrig2019end} for object recognition. Specifically, we first performed \(8\times\) downsampling on the NCars dataset through coordinate relocation. Subsequently, we employ different models to conduct \(2(4,8)\times\) super-resolution on the event stream and employ the object recognition method for identification. Table \ref{tab4} illustrates the results of object recognition comparison. We evaluate using accuracy (ACC) and area under the curve (AUC). GT signifies the utilization of results directly obtained from downsampled raw event streams. It can be observed that our method outperforms other approaches consistently across \(2(4,8)\times\) super-resolution scales. In comparison with previous methods, our approach achieves an average improvement of over 9\% in terms of ACC and AUC. These results demonstrate the superior detail restoration capability of our method.

\section{Conclusion}
In this paper, we introduced an efficient Recursive Multi-Branch Information Fusion Network (\method) for ESR tasks. \method leverages a carefully designed multi-branch network architecture, taking decoupled positive and negative events as well as coupled event frames as input to achieve super resolution of event streams. Additionally, we introduced attention-based Feature Fusion Module and Feature Exchange Module, which effectively integrate contextual information from neighboring event streams and facilitate the exchange of complementary information between positive and negative events. Furthermore, we explored the impact of data augmentation methods on ESR tasks and proposed an effective data augmentation strategy to enhance model robustness and performance. Results on both real and synthetic datasets demonstrated that our approach outperforms previous ESR methods across various metrics.

\section*{Acknowledgments}
This work was supported in part by the Guangxi Key R \& D Program (No. GuikeAB24010324), in part by the National Natural Science Foundation of China (No. 62088102, No. 62425101), in part by the Key-Area Research and Development Program of Guangdong Province (No. 2021B0101400002), and the Major Key Project of PCL (No. PCL2021A13).

\section*{Contribution Statement}
Quanmin Liang and Zhilin Huang made equal contributions. Kai Huang and Yonghong Tian are Corresponding Author. All the authors participated in designing research, analyzing data, and writing the paper.

\bibliographystyle{named}
\bibliography{RMFNet_maintext}

\end{document}